%% file: root.tex
\documentclass[letterpaper, 10 pt, conference]{ieeeconf}
\IEEEoverridecommandlockouts                
\usepackage{graphics} 
\usepackage{epsfig} 
\usepackage{mathptmx}
\usepackage{times} 
\usepackage{amsmath}
\usepackage{amssymb}
\usepackage{xcolor}
\usepackage{ulem}
\usepackage{tabularx,booktabs}
\usepackage{todonotes}
\usepackage{hyperref}
\newcolumntype{Y}{>{\centering\arraybackslash}X}
\newcolumntype{S}{>{\hsize=1.1\hsize}Y}

\title{\LARGE \bf
Contrastive Touch-to-Touch Pretraining}

\author{Samanta Rodriguez*, Yiming Dou*, William van den Bogert, \\ Miquel Oller, Kevin So, Andrew Owens, Nima Fazeli%
\thanks{*These authors contributed equally.}%
\thanks{$^{1}$University of Michigan
        {\tt\footnotesize {\{samanrod, ymdou, willvdb, oller, kvso, ahowens, nfz\}}@umich.edu}}%
\thanks{Supported by NSF GRFP \#2241144, NSF CAREER Awards \#2339071 and \#2337870, and NSF NRI \#2220876.}%
}
    
\begin{document}

\maketitle
\thispagestyle{empty}
\pagestyle{empty}

\begin{abstract}
     Today's tactile sensors have a variety of different designs, making it challenging to develop general-purpose methods for processing touch signals. In this paper, we learn a unified representation that captures the shared information between different tactile sensors. Unlike current approaches that focus on reconstruction or task-specific supervision, we leverage contrastive learning to integrate tactile signals from two different sensors into a shared embedding space, using a dataset in which the same objects are probed with multiple sensors. We apply this approach to paired touch signals from GelSlim and Soft Bubble sensors. We show that our learned features provide strong pretraining for downstream pose estimation and classification tasks. We also show that our embedding enables models trained using one touch sensor to be deployed using another without additional training. Project details can be found at \href{https://www.mmintlab.com/research/cttp/}{https://www.mmintlab.com/research/cttp/}.
\end{abstract}

\section{INTRODUCTION}

Tactile sensing is a fundamental enabler of dexterous robotic manipulation, providing critical information about contact forces, object properties, and interaction dynamics. However, unlike vision and audio, which have enjoyed significant standardization, tactile sensors vary greatly in their designs and sensing mechanisms. For instance, sensors such as GelSlim~\cite{gelsim_donlon}, Soft Bubble~\cite{softbub_tedrake}, and DIGIT~\cite{lambeta2020digit} each provide distinct forms of tactile feedback, from detailed surface deformations to broader contact geometries. While this diversity has enabled progress in specialized tasks, it also presents a major challenge: algorithms and models developed for one sensor type are rarely applicable to others. This sensor-specificity hampers the generalization of tactile manipulation techniques and necessitates labor-intensive adaptation efforts when working with new sensors, limiting their utility in real-world, unstructured environments.

Existing methods aimed at addressing these cross-sensor disparities, such as sensor calibration or transfer learning techniques, are often insufficient in practice. Calibration tends to be sensor-specific and difficult to generalize, while transfer learning struggles to overcome the substantial distribution shifts between sensors. Models trained on the data of one tactile sensor often fail to generalize effectively without significant retraining, as each sensor encodes its unique features based on its design. Although progress has been made in cross-sensor tactile modeling, existing approaches fail to capture the information shared between tactile sensors with different sensing modalities, leading to less than desirable performance in tasks such as object classification and pose estimation.

In this paper, we introduce Contrastive Touch-to-Touch Pretraining (CTTP), a self-supervised learning framework that uses paired tactile data to bridge the gap between different sensor modalities. Our approach is inspired by Contrastive Language-Image Pre-training (CLIP) \cite{clip} but extends the idea of contrastive learning to paired tactile data, which has not been explored in the context of visuo-tactile sensors. 
We obtain these paired tactile signals (from GelSlim and Soft Bubble sensors) from a robot grasping the same object in the same configuration but with different sensors. %
In this framework, paired data acts as positive examples, while unpaired data serves as negative examples, allowing us to discover representations that capture meaningful similarities between disparate sensors. Unlike self-supervised approaches that rely on reconstruction or task-specific objectives, our method focuses on preserving the shared information in the tactile signatures, enabling generalization across sensors. We show that this shared representation can be used to perform zero-shot classification and pose estimation across sensors, providing a scalable solution for integrating heterogeneous tactile sensors and opening new possibilities for tactile manipulation systems in real-world applications.

\begin{figure}
    \centering
    \includegraphics[width=\columnwidth]{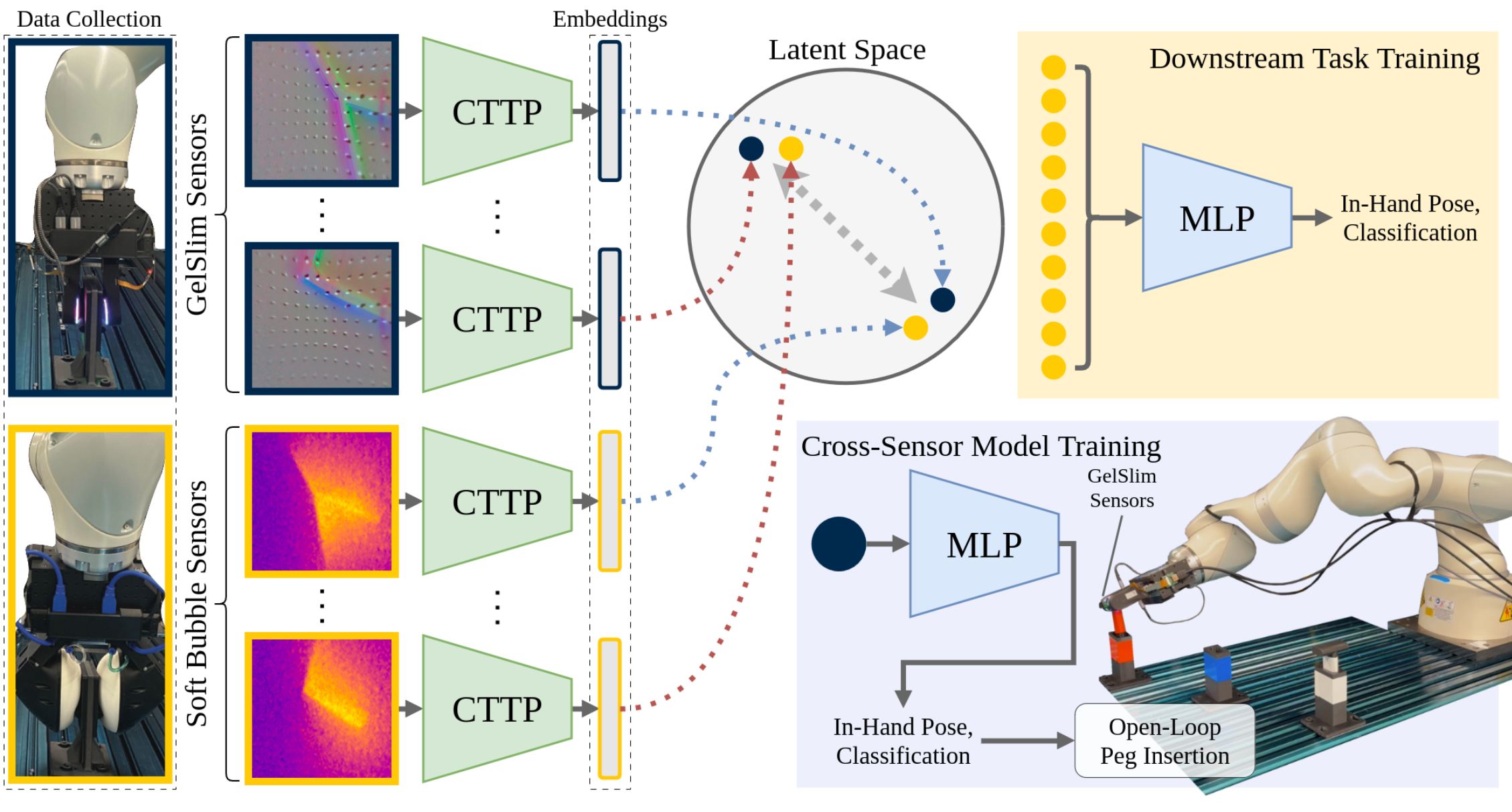}
    \caption{{\bf Contrastive Touch-to-touch Pretraining (CTTP).} We learn a joint embedding between signals from different tactile sensors. The resulting model learns a touch feature representation that conveys the physical properties of the touched object that are provided by both sensors, which is useful pretraining for downstream tasks. The embedding also enables ``zero shot'' transfer of downstream touch models from one sensor to another. } \vspace{-3mm}
    \label{fig:teaser}
\end{figure}

\section{RELATED WORKS}
\subsection{Representation Learning with Tactile Data} 
Prior works have been learning tactile representations using various methods, including aligning with other modalities, particularly vision~\cite{dou2024tactile,yang2024binding,cheng2024touch100k,yang2022touch,fu2024touch,tu2024texttoucher,yang2023generating,kerr2022self,lee2019making,lin2019learning}. Other work trains using task-specific supervised learning~\cite{gao2023objectfolder,T3} and masked autoencoding~\cite{T3}. In contrast, we pair touch signals from two different sensors, resulting in a representation that captures the information that is shared between them (rather than what is shared between vision and touch). This also allows us to directly transfer models between two touch sensors, which is not straightforward using these other approaches. %
Work from the audio community has successfully learned representations by performing contrastive learning on data processed in multiple formats~\cite{wang2021multi}, while we perform contrastive learning on data from two different sensors.
While recent work has shown that generative models can translate between tactile signals~\cite{rodriguez2024touch2touch}, it has not addressed the problem of learning touch representations using these paired signals.

\subsection{Vision-Based Tactile Sensors}
To capture touch signals, vision-based tactile sensors use cameras that detect deformations of an elastomer layer installed on the top of it. During recent years, many vision-based tactile sensors have been introduced, including GelSight~\cite{yuan2017gelsight}, GelSlim~\cite{gelsim_donlon,taylor2022gelslim}, Soft Bubble~\cite{kuppuswamy2020soft,softbub_tedrake}, DIGIT~\cite{digit_tactile_sensor}, Finger Vision~\cite{fingerVision}, and DenseTact~\cite{Do2022DenseTactOT}. The touch signals are represented as either 2D images or 3D depth maps. 
We use Soft Bubble and GelSlim sensors in this work. 
The Soft Bubble sensor uses a camera-based depth sensor to capture the deformation of a thin, compliant, and air-filled membrane when it is warped by external contacts. 
The GelSlim sensor uses an RGB camera to measure deformations of an elastomer. The elastomer is illuminated by LED lights, and changes in deformations correspond to changes in the material's color.
We perform pretraining on these two sensors to show that our CTTP learns useful representations even on sensors that are largely different in deformations, compliance, contact areas, and representation.

\subsection{Tasks and Algorithms for Tactile Sensing} 
Recent works have showcased the usefulness of vision-based tactile sensors on various tasks, including material estimation~\cite{dou2024tactile,gao2023objectfolder,yang2024binding}, grasping stability prediction~\cite{gao2023objectfolder,yang2024binding}, in-hand object pose estimation (e.g., Soft Bubble~\cite{kuppuswamy2019fast}, GelSlim~\cite{kim2022active}, and DIGIT~\cite{suresh2023midastouch}), local geometry estimation (e.g., Soft Bubble~\cite{kuppuswamy2019fast}, GelSlim~\cite{taylor2022gelslim}, GelSight~\cite{gao2023objectfolder}, and DIGIT~\cite{xu2023visual}), and force field estimation (e.g., Soft Bubble~\cite{kuppuswamy2020soft}, GelSlim~\cite{taylor2022gelslim}, Finger Vision~\cite{yamaguchi2016combining}). These methods further improve success on manipulation tasks, such as peg-in-hole insertion~\cite{kim2022active}, in-hand pivoting~\cite{oller2023manipulation}, and dense packing~\cite{ai2024robopack}.
However, all of these methods are tied to specific tactile sensors.
UniTouch~\cite{yang2024binding} and T3 model~\cite{T3} tackle this problem by designing a unified model for tactile sensors, but they require either aligned vision data or an intensive labelling process (e.g., object class and pose). Our approach, by contrast, leverages aligned tactile data and does not require additional labeling.

\input{text/3-method}
\input{text/4-results}
\input{text/5-discussion}

\bibliographystyle{IEEEtran}
\bibliography{references}

\end{document}

%% file: text/3-method.tex
\section{METHOD}

In this section, we present the details of our approach for learning a shared representation across tactile sensors using contrastive self-supervised learning. First, we describe how we adapt contrastive learning, which is typically applied in vision tasks, to train on paired tactile data. Next, we explain the role of the batch size selection for CTTP performance. Finally, we describe the experimental setup, including the robot and tactile sensors used, and the process for pairing tactile data across different sensors.

\subsection{Contrastive Learning on Paired Tactile Data}

Contrastive learning has emerged as a powerful tool for self-supervised representation learning across various domains, including vision, language, and multimodal tasks. In a typical contrastive learning setup, the goal is to learn an embedding space where similar inputs (positive pairs) are pulled together and dissimilar inputs (negative pairs) are pushed apart. For instance, in vision-language models like CLIP~\cite{clip}, images and text captions are aligned based on whether they refer to the same scene. Inspired by this cross-modal alignment, we adopt a similar approach for tactile data, where we aim to align different tactile sensors that capture the same object in the same configuration into a shared latent space, treating these as positive pairs and pushing apart other signatures.

More formally, for each object we collect tactile signals from two sensors, and define a positive pair as $(\mathbf{t}_1, \mathbf{t}_2)$, where $\mathbf{t}_1$ and $\mathbf{t}_2$ are the tactile signals from the two sensors when grasping the same object in the same configuration. Negative pairs consist of tactile signals from either sensor that come from different objects or distinct ways of grasping the same object. Let $\mathbf{z}_i^{(1)}$ and $\mathbf{z}_i^{(2)}$ represent the latent embeddings of these tactile signals. The objective of contrastive learning is to learn an embedding function $f_\theta$ that minimizes the distance of latent vectors between positive pairs while maximizing the distance between negative pairs. Fig.\ref{fig:teaser} shows a schematic of our method.

We use the InfoNCE loss~\cite{chen2020simple}, which is commonly used in contrastive learning, to encourage the separation of positive and negative pairs in the embedding space. For each pair of tactile signals, the loss is computed as:

\begin{align}
    \mathcal{L}_{\text{NCE}} = - \log \frac{\exp \left( \text{sim}(\mathbf{z}_i^{(1)}, \mathbf{z}_i^{(2)}) / \tau \right)}{\sum_{j=1}^{N} \exp \left( \text{sim}(\mathbf{z}_i^{(1)}, \mathbf{z}_j^{(1\setminus2)}) / \tau \right)},
\end{align}
where $\text{sim}(\cdot, \cdot)$ is the cosine similarity between embeddings:
\begin{align}
\text{sim}(\mathbf{z}_i, \mathbf{z}_j) = \frac{\mathbf{z}_i \cdot \mathbf{z}_j}{\|\mathbf{z}_i\| \|\mathbf{z}_j\|},
\end{align}
and $\tau$ is a temperature parameter that controls the sharpness of the distribution, and $N$ is the number of negative pairs in the batch. By optimizing this loss, we maximize the similarity between the latent representations of positive pairs $(\mathbf{z}_1, \mathbf{z}_2)$, while minimizing the similarity between the anchor $\mathbf{z}_1$ and the negative examples $\{\mathbf{z}_n\}$.

\subsection{Experimental Setup}
Our experiments are conducted using a robot equipped with two pairs of tactile sensors: GelSlim and Soft Bubble sensors. We observe the robot setup with these tactile sensors and corresponding tactile signals in Fig.\ref{fig:teaser}. We collect data from a diverse set of objects, varying in shape and size. Each object is placed in a fixed position within the robot’s workspace, and the robot is instructed to grasp the object at different configurations, capturing both tactile signatures from each sensor, as well as the robot’s pose information.

Data collection is performed independently for each tactile sensor, and tactile signatures are paired when the same object is grasped in the same configuration by different sensors. This allows us to establish positive pairs for contrastive learning. For negative pairs, we collect data from different objects or significantly varied grasp configurations of the same object. We ensure the dataset captures a wide range of tactile signatures across objects and grasp poses to support generalization of our model across varying sensors and novel objects.

\subsection{Default Setting}
We use ResNet-50~\cite{he2016deep} as the encoder network, and a two layer MLP projection head to project the representation to a 64-dimensional latent space. We optimize the model using Adam with a learning rate of $3\times10^{-4}$, and we train at batch size 128 for 100 epochs.

%% file: text/4-results.tex
\pdfminorversion 4
\section{Experiments and Results}\label{section:results}
We evaluate CTTP on two main tasks: (i) tool classification and (ii) in-hand pose estimation. We benchmark CTTP's performance by testing it against several baselines:
\begin{itemize}
    \item \textbf{ImageNet PT}: ResNet-50 \cite{resnet} pretrained on the ImageNet \cite{imagenet} dataset.
    \item \textbf{ResNet RI}: ResNet-50 with random initialization.
    \item \textbf{T3 PT}: The T3 \cite{T3} architecture pretrained on reconstruction.
    \item \textbf{T3 Class PT}: The T3 architecture pretrained on object classification.
    \item \textbf{T3 Pose PT}: The T3 architecture pretrained on in-hand pose estimation.
\end{itemize}

We emphasize that the final two baselines require labeled data to learn the latent representation of tactile information, while CTTP does not. For the T3 baselines requiring labeled data, we evaluate the latent space learned from the labels on other downstream tasks to keep evaluation fair. Both of our downstream tasks require learning predictions from latent spaces using neural networks. As such, we examine two training and evaluation datasets for each downstream task:
\begin{itemize}
    \item \textbf{Unseen Grasps:} We train the classification and in-hand pose estimation networks on the latent vectors that represent the original training dataset used to learn the representations, then evaluate the tasks on latent vectors that represent grasps which are not seen during the representation learning nor the downstream task learning. These grasps are on the same tools used for representation learning.
    \item \textbf{Unseen Tools:} We train the classification and in-hand pose estimation networks on the latent vectors that represent grasps on tools which were not used for representation learning, then evaluate the tasks on a separate set of latent vectors that represent grasps which are not seen during the representation learning nor the downstream task learning. These grasps are on the same tools used for the downstream task learning.
\end{itemize}

Additionally, we examine two generalization cases:
\begin{itemize}
    \item \textbf{Generalization Within Sensor:} In this case, the downstream tasks are trained on latent vectors that represent one sensor's images and then tested using unseen images from that same sensor.
    \item \textbf{Generalization Across Sensors:} In this case, downstream tasks are trained on latent vectors that represent one sensor's images, then tested using images from the other sensor. We emphasize that CTTP's high performance in this case when compared to the other methods is due to the alignment of the two sensors' latent representations.
\end{itemize}

\subsection{Classification}\label{subsection:class_task}
\begin{figure}
    \centering
    \includegraphics[width=\columnwidth]{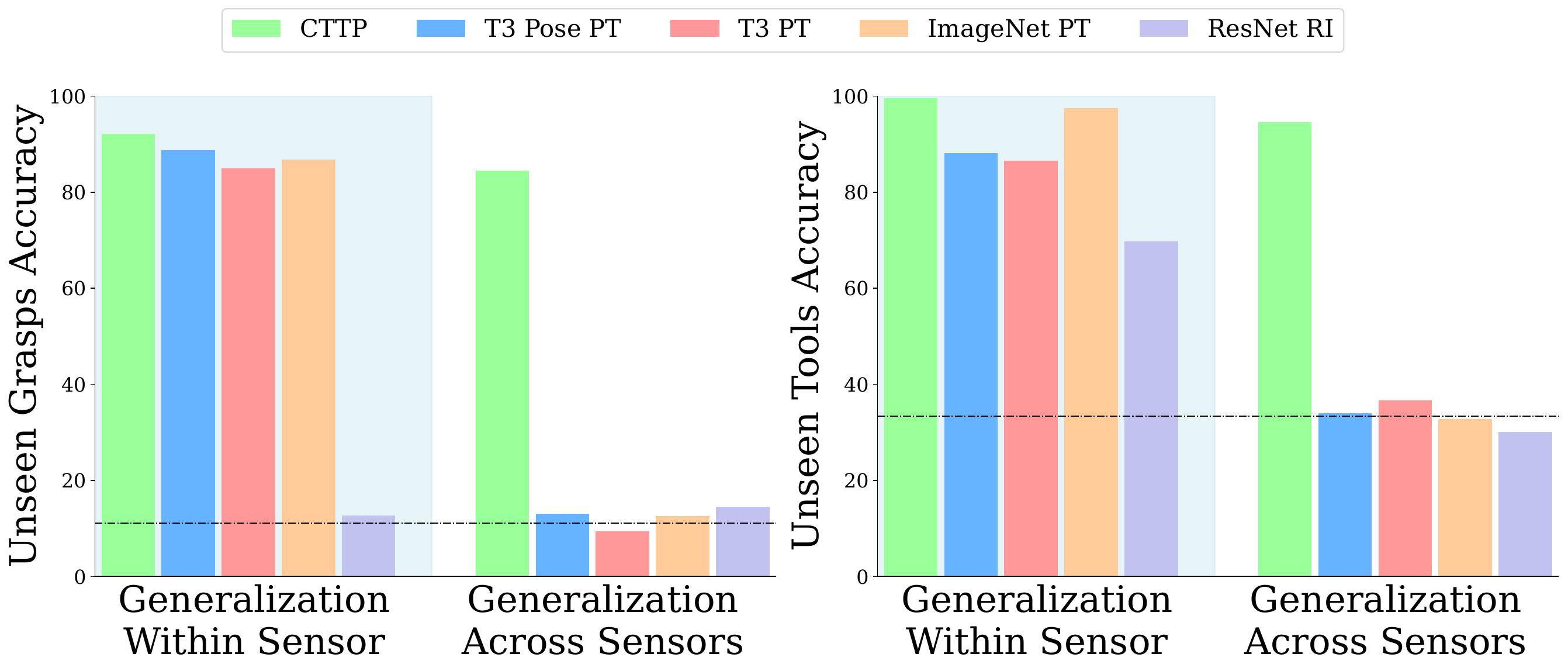}
    \caption{\textbf{Representation Learning Models Comparison on Classification Accuracy}. We compare CTTP to our baselines on the downstream task of tool classification. We evaluate their performance in three areas: generalization within a single visuo-tactile sensor, generalization across different visuo-tactile sensors, and generalization to unseen tools. For reference, the dotted line represents random chance performance.} 
    \label{fig:class_models_comparison}
\end{figure}

\begin{figure}
    \centering
    \includegraphics[width=\columnwidth]{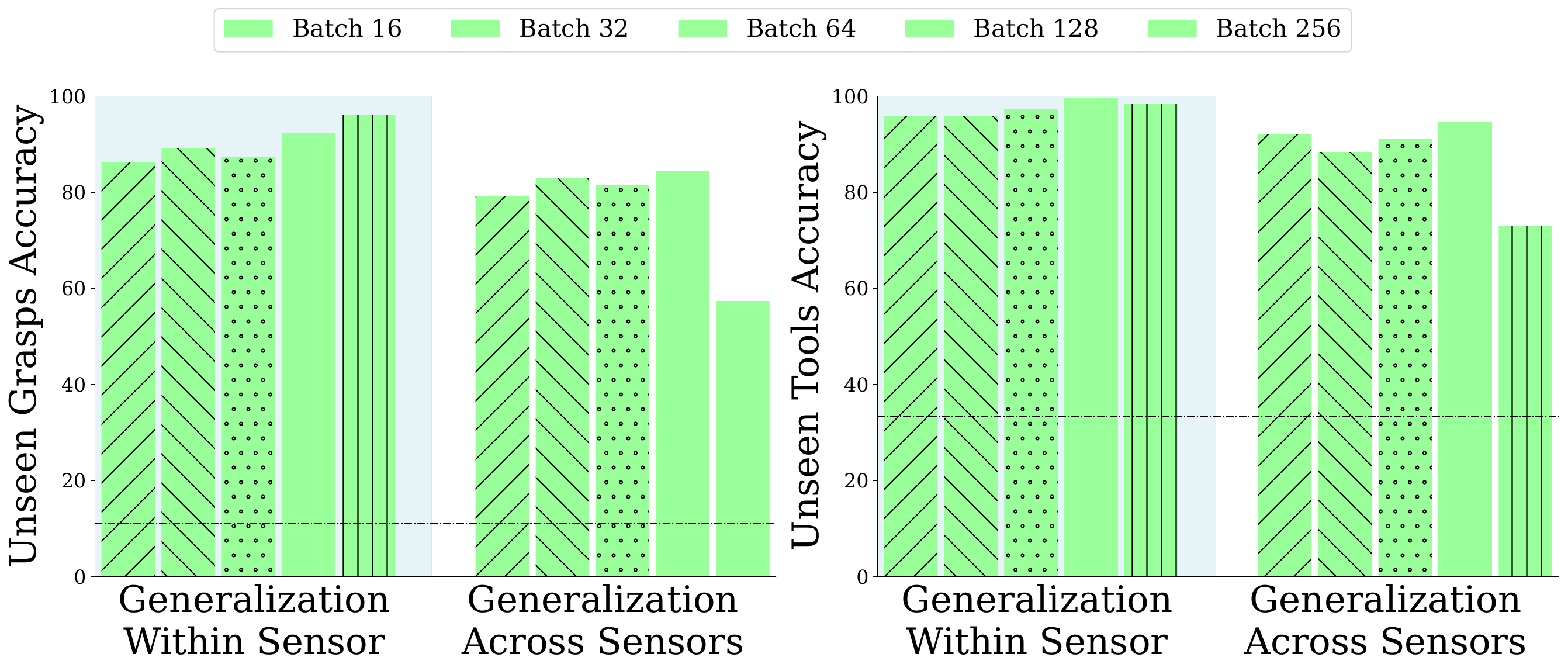}
    \caption{\textbf{CTTP Batch Size Comparison on Classification Accuracy}. We compare CTTP trained on different batch sizes on the downstream task of tool classification. We evaluate their performance in three areas: generalization within a single visuo-tactile sensor, generalization across different visuo-tactile sensors, and generalization to unseen tools. For reference, the dotted line represents random chance performance.}
    \label{fig:class_batches_comparison}
\end{figure}

Here, we evaluate our method on the downstream task of tool classification. We train a single linear layer on the embeddings generated by each of our baselines and classify within 9 classes (9 different tools) for unseen grasps and 3 classes (3 different tools) for unseen tools. We train the model by minimizing the cross-entropy loss between the ground truth labels and the predicted class probabilities. For evaluation, we utilize top-1 accuracy by selecting the class with the highest predicted probability. Fig.~\ref{fig:class_models_comparison} shows the results of CTTP in comparison to the baselines, and Fig.~\ref{fig:class_batches_comparison} displays the tool classification results for CTTP trained with different batch sizes. 

In Fig. \ref{fig:class_models_comparison}, we observe that, with the exception of ResNet RI, all methods achieve an accuracy above 80\%, with CTTP achieving the highest accuracy. However, when it comes to generalization across tactile sensors, our results show a significant drop in performance for all baselines, approximating random chance (denoted by a dotted line). In contrast, CTTP maintains significantly higher performance. Additionally, we observe that CTTP and ImageNet PT surpass both T3 baselines when evaluated on the unseen tools, suggesting higher tool generalization performance.

Figure \ref{fig:class_batches_comparison} shows how training with different batch sizes impacts CTTP performance. Our final CTTP model uses a batch size of 128. In Fig. \ref{fig:class_batches_comparison}, we observe that increasing the batch size improves classification accuracy, particularly for generalization within the same sensor. However, performance significantly drops for generalization across different sensors when the batch size is increased to 256. When evaluating the impact of batch size on generalization to unseen tools, we observe a trend similar to that of generalization to unseen grasps.

\subsection{In-Hand Pose Estimation}\label{subsection:ihp_task}
We test the latent space generated by CTTP against the baselines for the task of estimating the SE(2) in-hand pose of a grasped object. For all results, a fully-connected network (two 256-dimensional hidden layers) takes each 2048-dimensional latent vector as input and outputs a 3-dimensional in-hand pose $(y,z,\theta)$. The parameters of this network are learned by minimizing the MSE between the estimated and ground-truth in-hand pose. Two separate predictors were trained to estimate the SE(2) pose for each finger in the parallel plate gripper.

The unseen grasp results for in-hand pose estimation can be seen in Table \ref{tab:ihp_unseen_grasps}. When generalizing within the latent vector-represented sensor, the T3 and ImageNet PT baselines exhibit similar error magnitudes while our method outperforms them most significantly in the $\theta$ orientation error, with the majority of errors falling within $\pm 1^\circ$. The randomly initialized ResNet performed the worst in this case. When generalizing across sensors, our method outperforms the other methods in $(y,z)$ translation error and again most significantly in the $\theta$ orientation error, with the majority of errors falling within $\pm 5^\circ$. The self-supervised T3 PT baseline performed the worst in this case.

The unseen tools results for in-hand pose estimation can be seen in Table \ref{tab:ihp_unseen_tools}. When generalizing within the latent vector-represented sensor, again, the T3 and ImageNet PT baselines exhibit similar error margins, and our method outperforms them in the $\theta$ orientation error. In terms of translation error, testing on unseen tools appears to enhance the performance advantage of CTTP over other methods, compared to testing on unseen grasps. The randomly initialized ResNet again performed the worst. When generalizing across sensors, the CTTP latent vectors predict the majority of orientation errors to be within $\pm 5^\circ$, far exceeding the other methods. In translation, our method maintains the majority of errors to be within $\pm 3$ mm, indicating some ambiguity while still outperforming all other methods by at least $1$ mm. The self-supervised T3 PT baseline again performed the worst in this case. The second-best latent representation for in-hand pose estimation in this case is the T3 Class PT baseline. Thus, this is the baseline with which we compared the CTTP latent representation for in-hand pose estimation in the use case of robotic peg insertion in Section \ref{subsection:insertion_task}.

\begin{table}[h]
\begin{center}
\caption{In-Hand Pose Estimation Errors for Unseen Grasps}
\label{tab:ihp_unseen_grasps}
\begin{tabularx}{\columnwidth}{r *{3}{Y}}\toprule
\textbf{Method} & $y$ Error (mm) & $z$ Error (mm) & $\theta$ Error ($^{\circ}$) \\
\midrule
\multicolumn{4}{c}{\textbf{Generalization Within Sensor}} \\
\midrule
CTTP & $\mathbf{-0.01 \pm 0.21}$ & $\mathbf{0.02 \pm 0.31}$ & $\mathbf{0.03 \pm 0.56}$  \\
T3 Class PT & $-0.01 \pm 0.73$ & $0.01 \pm 0.71$ & $-0.04 \pm 2.81$  \\
T3 PT & $-0.0 \pm 0.42$ & $0.02 \pm 0.38$ & $-0.14 \pm 2.42$  \\
ImageNet PT & $-0.0 \pm 0.72$ & $0.01 \pm 0.72$ & $-0.26 \pm 3.51$  \\
ResNet RI & $-0.12 \pm 1.41$ & $0.06 \pm 1.18$ & $-1.69 \pm 10.47$  \\
\midrule
\multicolumn{4}{c}{\textbf{Generalization Across Sensors}} \\
\midrule
CTTP & $\mathbf{-0.03 \pm 1.79}$ & $\mathbf{0.12 \pm 3.32}$ & $\mathbf{0.25 \pm 4.39}$  \\
T3 Class PT & $-0.25 \pm 4.38$ & $0.82 \pm 3.6$ & $-4.22 \pm 18.78$  \\
T3 PT & $-10.26 \pm 15.81$ & $-4.66 \pm 5.34$ & $-33.87 \pm 38.87$  \\
ImageNet PT & $-4.67 \pm 6.15$ & $-5.2 \pm 4.97$ & $5.85 \pm 23.64$  \\
ResNet RI & $-0.36 \pm 3.16$ & $-4.9 \pm 4.97$ & $-2.8 \pm 15.27$  \\
\bottomrule
\end{tabularx}
\end{center}
\end{table}

\begin{table}[h]
\begin{center}
\caption{In-Hand Pose Estimation Errors for Unseen Tools}
\label{tab:ihp_unseen_tools}
\begin{tabularx}{\columnwidth}{r *{3}{Y}}\toprule
\textbf{Method} & $y$ Error (mm) & $z$ Error (mm) & $\theta$ Error ($^{\circ}$) \\
\midrule
\multicolumn{4}{c}{\textbf{Generalization Within Sensor}} \\
\midrule
CTTP & $\mathbf{0.01 \pm 0.24}$ & $\mathbf{0.0 \pm 0.09}$ & $\mathbf{-0.01 \pm 0.34}$  \\
T3 Class PT & $0.04 \pm 1.02$ & $0.01 \pm 0.65$ & $-0.1 \pm 3.13$  \\
T3 PT & $-0.02 \pm 0.59$ & $-0.01 \pm 0.28$ & $-0.03 \pm 2.34$  \\
ImageNet PT & $-0.02 \pm 0.86$ & $0.01 \pm 0.59$ & $-0.1 \pm 2.5$  \\
ResNet RI & $0.08 \pm 1.2$ & $0.04 \pm 0.57$ & $0.23 \pm 5.85$  \\
\midrule
\multicolumn{4}{c}{\textbf{Generalization Across Sensors}} \\
\midrule
CTTP & $\mathbf{0.23 \pm 2.79}$ & $\mathbf{-0.17 \pm 2.41}$ & $\mathbf{-0.01 \pm 4.39}$  \\
T3 Class PT & $-0.95 \pm 5.71$ & $0.45 \pm 3.57$ & $-2.67 \pm 18.47$  \\
T3 PT & $1.8 \pm 14.12$ & $-2.52 \pm 5.76$ & $19.04 \pm 27.54$  \\
ImageNet PT & $0.3 \pm 3.34$ & $-2.05 \pm 3.44$ & $-1.84 \pm 21.82$  \\
ResNet RI & $2.1 \pm 5.35$ & $-5.27 \pm 6.83$ & $1.68 \pm 22.79$  \\
\bottomrule
\end{tabularx}
\end{center}
\end{table}

\subsection{Latent Space Visualization}\label{subsection:tsne_experiment}
\begin{figure*}
    \centering
    \includegraphics[width=0.9\linewidth]{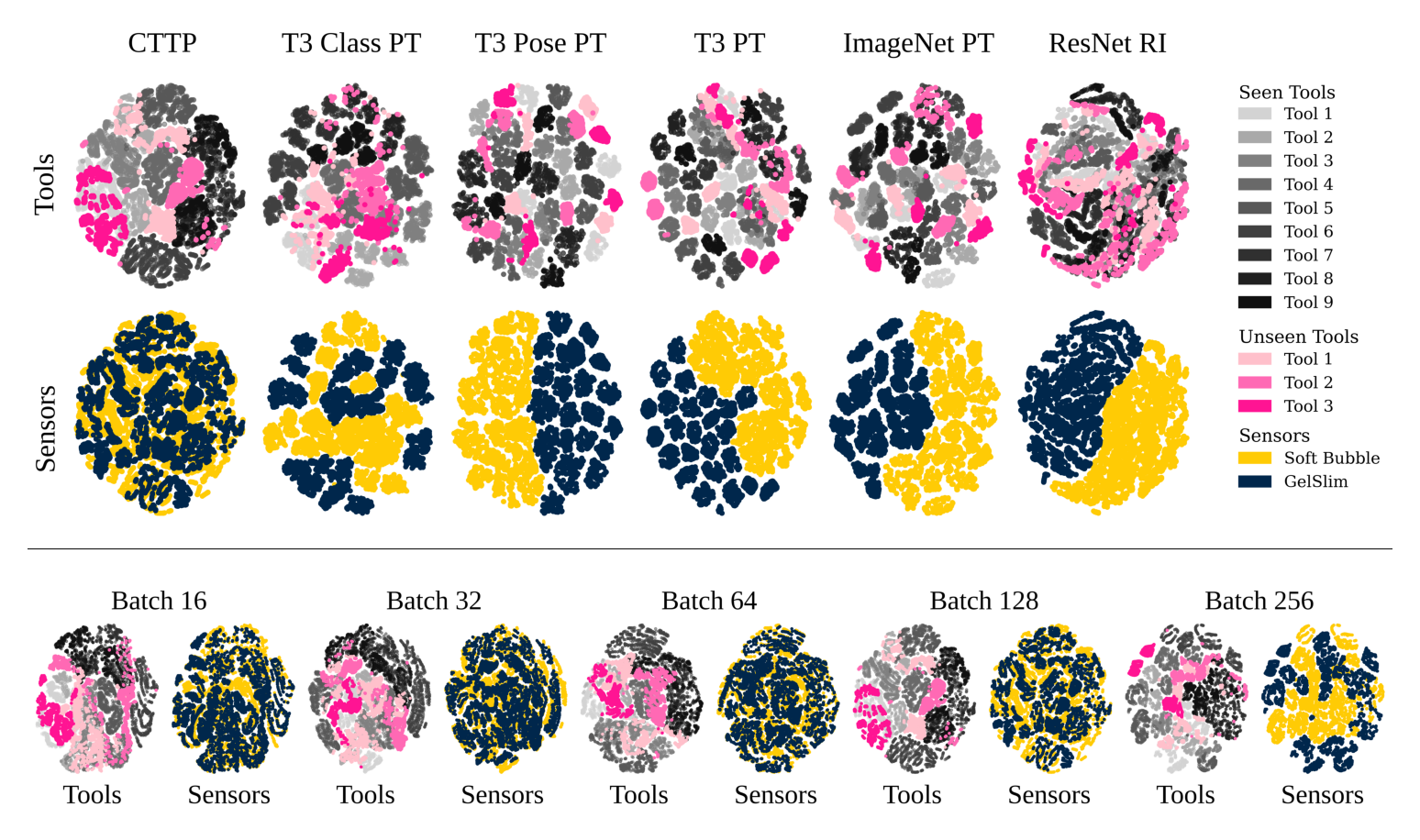}
    \caption{\textbf{TSNE Comparison.} we present the results of a t-SNE analysis on embeddings from both seen and unseen tools in our dataset. We conduct this analysis for CTTP, several baselines models (top), and CTTP trained with different batch sizes (bottom). For this analysis, we focus on visualizing the structure and relationships between the embeddings, focusing on tool differentiation (gray and pink) and sensor alignment (maize and blue). We consider the model successful when the t-SNE shows groupings of tools and, at the same time, the sensor colors overlap (sensors are aligned). Our CTTP model is trained on a batch size of 128 (bottom).}
    \label{fig:tsne_comparison}
\end{figure*}

We used t-SNE (t-distributed Stochastic Neighbor Embedding) \cite{tsne} to map the 2048-dimensional latent vector generated by CTTP to a low-dimensional 2D space. In Fig. \ref{fig:tsne_comparison}, we present the results of a t-SNE analysis on embeddings from both seen and unseen tools in our dataset. We conduct this analysis for CTTP, several baseline models, and CTTP trained with different batch sizes. This visualization aims to illustrate the structure and relationships between the embeddings.

To assess each model's success, we look for two key features in the t-SNE plot: tool differentiation (represented by gray and pink) and sensor alignment (represented by maize and blue). Tool differentiation means that the plot shows distinct tool groupings, indicating that the model correctly differentiates between different tools. Sensor alignment, on the other hand, shows overlapping colors for sensors, indicating that the model effectively aligns the sensors and that the latent space is sensor agnostic.

A successful model will display clear clusters for each tool and overlapping sensor colors within each cluster, showing that tools are distinguished and the sensor data is aligned.

The top of Fig.\ref{fig:tsne_comparison} shows the t-SNE representations comparing CTTP and our baselines. We observe that, with the exception of ResNet RI, all models are able to differentiate between tools, with CTTP showing larger and more unified clusters per tool. For sensor distribution, the t-SNE projection shows that the baselines maintain a separation between Soft Bubble and GelSlim regions. In contrast, CTTP has high overlaps between these regions. This outcome is desirable because our goal is to align the sensors and create an embedding space that clearly differentiates tools while being agnostic to sensor-specific differences.

The bottom of Fig.\ref{fig:tsne_comparison} shows how the embedding space of CTTP changes with batch size selection. We can see how, as we increase the batch size, the space differentiates better between tool classes but simultaneously the sensors lose alignment. 

\subsection{Insertion Task}\label{subsection:insertion_task}
\begin{figure}
    \centering
    \includegraphics[width=\columnwidth]{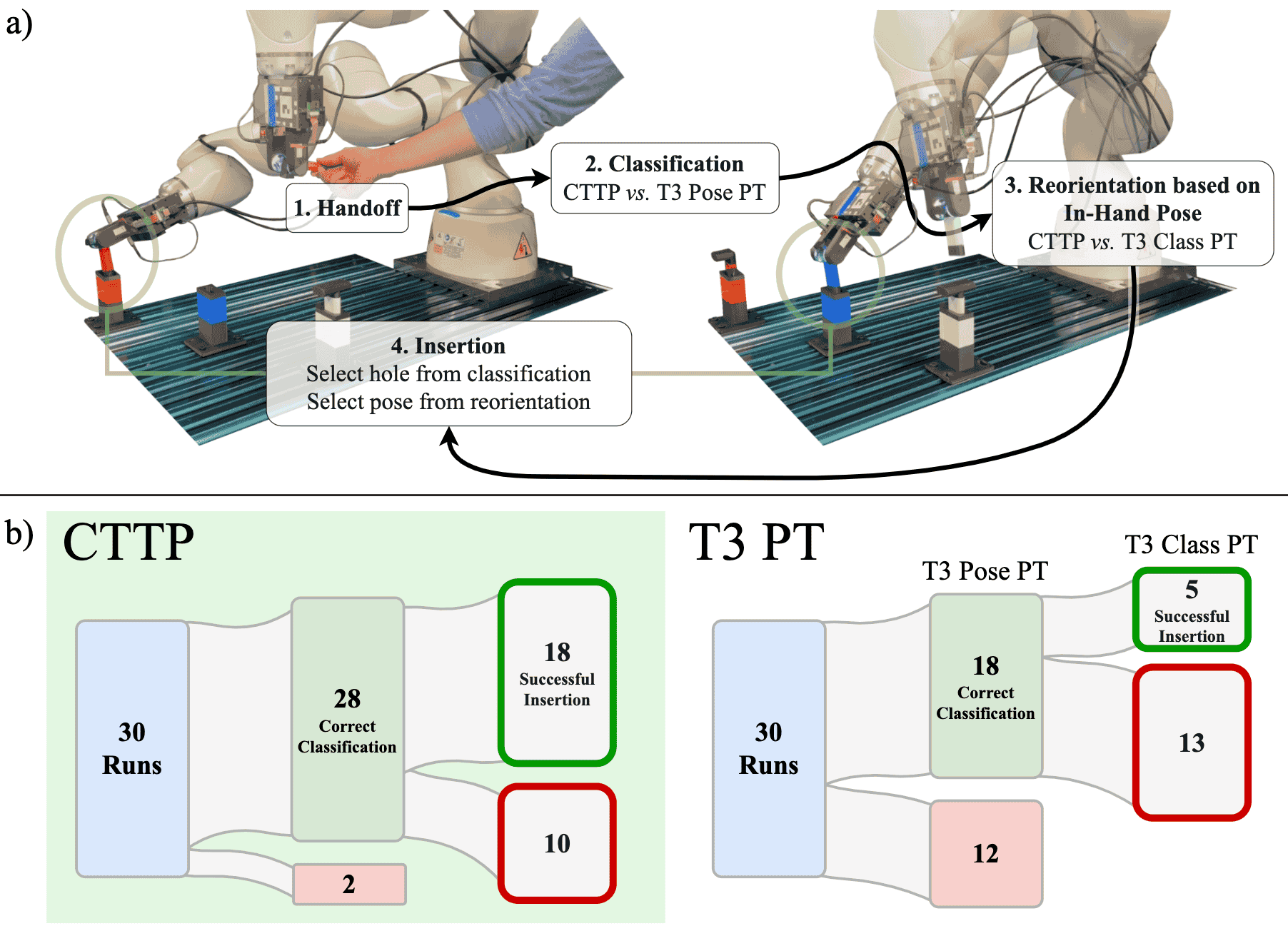}
    \caption{{\bf Insertion Task.} We use our representation to perform peg insertion tasks using both tool classification and in-hand pose estimation. a) Our testing setup using each of the three unseen tools and corresponding holes. After handing the tool to the robot, classification occurs which allows the robot to select the correct hole. b) The CTTP-generated latent space performs far better in cross-sensor task transfer.}
    \label{fig:insertion_task}
\end{figure}

We leverage our classification and in-hand pose estimation models to perform a peg insertion task, and compare the performance of CTTP-generated latent spaces against those generated by T3 for this purpose. The robot setup is shown in Fig. \ref{fig:insertion_task}. The insertion task consists of four stages: handoff, classification, reorientation, and insertion. First, the robot is handed an unknown tool at an unknown pose. Second, using the classification model discussed in Section \ref{subsection:class_task}, the robot identifies the tool, and selects the correct hole for peg insertion. Third, the robot determines the in-hand pose of the tool and reorients it to align with the hole, preparing for placement. Finally, the robot inserts the tool into the corresponding hole. 

For this insertion task, we are interested in cross-sensor generalization, and thus test the class and in-hand pose predictors that were trained on Soft Bubble latent vectors, while  our testing setup uses GelSlims. In this experiment, we compare the performance of CTTP and T3, and use T3 Pose PT as the pretraining model for the classification stage and T3 Class PT as the pretraining model for the pose estimation stage. This ensures a fair comparison in which the task had not been seen during pretraining. We performed these experiments with three unseen tools that were not used during pretraining. The results of these real robot experiments are shown in Fig. \ref{fig:insertion_task}, showing that our CTTP-derived latent vectors lead to a much higher insertion success rate (18) than the second-best representations from T3 (5). In online classification, the CTTP-derived latent space correctly identified the tool 28/30 times, compared to the T3 Pose PT latent space correctly identifying the tool only 18/30 times.

%% file: text/5-discussion.tex
\section{Discussion}
In this paper, we showed that we can obtain useful tactile representations by learning a joint embedding between the signals of different touch sensors. This representation performs well on downstream touch understanding tasks, and it allows us to transfer touch models trained on one sensor to another, in contrast to other tactile embedding methods lacking this ability. Our work opens two new directions. The first is studying the merits of cross-modal touch supervision, and how it relates to other supervisory signals, such as vision and touch. The second is developing new contrastive learning models and augmentation strategies.

\noindent {\bf Limitations.} We conducted our experiments on two types of tactile sensors, GelSlim and Soft Bubble sensors, as they are commonly used in robotics and have very different form factors. Further experiments have yet to reveal whether aligning the latent spaces of more than two tactile sensors would improve CTTP generalization due to increased dataset size.

\noindent {\bf Extrapolations.} Fig. \ref{fig:tsne_comparison} shows CTTP to be the only method where latent vectors from both sensors are thoroughly mixed throughout the latent space, a finding which agrees with the performance of CTTP on cross-sensor model training and downstream tasks we show in Section \ref{section:results}. Additionally, increasing the batch size of CTTP to 256 causes the sensors to lose alignment in the latent space. This is likely because the batch size controls the ratio of negative to positive samples in the contrastive learning process. The number of negative samples increases with batch size while the positive samples remain the same. 
Thus, we recommend a batch size of 128 for training CTTP.

It seems all methods generate a latent representation which performs similarly on downstream tasks, regardless if that task is learned on the tools which trained the latent representation or not. The unseen tools and unseen grasps train/test dataset configurations yielded similar results in both classification and in-hand pose estimation. Additionally, while our method does outperform others in within-sensor generalization, the real strength of CTTP is in across-sensor generalization. No other method exceeded random choice for across-sensor classification. In across-sensor in-hand pose estimation, CTTP is the only method which yields a usable range of orientation errors. We have shown that a pretrained CTTP model provides the means to utilize a labeled dataset from one tactile sensor to train downstream tasks that enable robotic manipulation with an entirely different tactile sensor. This could potentially bridge gaps in the tactile sensing community. As increasing diversity in available tactile sensing in turn increases diversity in datasets and algorithms, CTTP can help ensure that progress made in one tactile laboratory is progress made in another.

%% file: root.bbl
\begin{thebibliography}{10}
\providecommand{\url}[1]{#1}
\csname url@samestyle\endcsname
\providecommand{\newblock}{\relax}
\providecommand{\bibinfo}[2]{#2}
\providecommand{\BIBentrySTDinterwordspacing}{\spaceskip=0pt\relax}
\providecommand{\BIBentryALTinterwordstretchfactor}{4}
\providecommand{\BIBentryALTinterwordspacing}{\spaceskip=\fontdimen2\font plus
\BIBentryALTinterwordstretchfactor\fontdimen3\font minus \fontdimen4\font\relax}
\providecommand{\BIBforeignlanguage}[2]{{%
\expandafter\ifx\csname l@#1\endcsname\relax
\typeout{** WARNING: IEEEtran.bst: No hyphenation pattern has been}%
\typeout{** loaded for the language `#1'. Using the pattern for}%
\typeout{** the default language instead.}%
\else
\language=\csname l@#1\endcsname
\fi
#2}}
\providecommand{\BIBdecl}{\relax}
\BIBdecl

\bibitem{gelsim_donlon}
\BIBentryALTinterwordspacing
E.~Donlon, S.~Dong, M.~Liu, J.~Li, E.~H. Adelson, and A.~Rodriguez, ``Gelslim: {A} high-resolution, compact, robust, and calibrated tactile-sensing finger,'' \emph{CoRR}, vol. abs/1803.00628, 2018. [Online]. Available: \url{http://arxiv.org/abs/1803.00628}
\BIBentrySTDinterwordspacing

\bibitem{softbub_tedrake}
\BIBentryALTinterwordspacing
A.~Alspach, K.~Hashimoto, N.~Kuppuswamy, and R.~Tedrake, ``Soft-bubble: {A} highly compliant dense geometry tactile sensor for robot manipulation,'' \emph{2019 2nd IEEE International Conference on Soft Robotics (RoboSoft)}, pp. 597--604, 2019. [Online]. Available: \url{http://arxiv.org/abs/1904.02252}
\BIBentrySTDinterwordspacing

\bibitem{lambeta2020digit}
M.~Lambeta, P.-W. Chou, S.~Tian, B.~Yang, B.~Maloon, V.~R. Most, D.~Stroud, R.~Santos, A.~Byagowi, G.~Kammerer \emph{et~al.}, ``Digit: A novel design for a low-cost compact high-resolution tactile sensor with application to in-hand manipulation,'' \emph{IEEE Robotics and Automation Letters}, vol.~5, no.~3, pp. 3838--3845, 2020.

\bibitem{clip}
A.~Radford, J.~W. Kim, C.~Hallacy, A.~Ramesh, G.~Goh, S.~Agarwal, G.~Sastry, A.~Askell, P.~Mishkin, J.~Clark \emph{et~al.}, ``Learning transferable visual models from natural language supervision,'' pp. 8748--8763, 2021.

\bibitem{dou2024tactile}
Y.~Dou, F.~Yang, Y.~Liu, A.~Loquercio, and A.~Owens, ``Tactile-augmented radiance fields,'' \emph{arXiv preprint arXiv:2405.04534}, 2024.

\bibitem{yang2024binding}
F.~Yang, C.~Feng, Z.~Chen, H.~Park, D.~Wang, Y.~Dou, Z.~Zeng, X.~Chen, R.~Gangopadhyay, A.~Owens \emph{et~al.}, ``Binding touch to everything: Learning unified multimodal tactile representations,'' \emph{arXiv preprint arXiv:2401.18084}, 2024.

\bibitem{cheng2024touch100k}
N.~Cheng, C.~Guan, J.~Gao, W.~Wang, Y.~Li, F.~Meng, J.~Zhou, B.~Fang, J.~Xu, and W.~Han, ``Touch100k: A large-scale touch-language-vision dataset for touch-centric multimodal representation,'' \emph{arXiv preprint arXiv:2406.03813}, 2024.

\bibitem{yang2022touch}
F.~Yang, C.~Ma, J.~Zhang, J.~Zhu, W.~Yuan, and A.~Owens, ``Touch and go: Learning from human-collected vision and touch,'' \emph{arXiv preprint arXiv:2211.12498}, 2022.

\bibitem{fu2024touch}
L.~Fu, G.~Datta, H.~Huang, W.~C.-H. Panitch, J.~Drake, J.~Ortiz, M.~Mukadam, M.~Lambeta, R.~Calandra, and K.~Goldberg, ``A touch, vision, and language dataset for multimodal alignment,'' \emph{arXiv preprint arXiv:2402.13232}, 2024.

\bibitem{tu2024texttoucher}
J.~Tu, H.~Fu, F.~Yang, H.~Zhao, C.~Zhang, and H.~Qian, ``Texttoucher: Fine-grained text-to-touch generation,'' \emph{arXiv preprint arXiv:2409.05427}, 2024.

\bibitem{yang2023generating}
F.~Yang, J.~Zhang, and A.~Owens, ``Generating visual scenes from touch,'' in \emph{Proceedings of the IEEE/CVF International Conference on Computer Vision}, 2023, pp. 22\,070--22\,080.

\bibitem{kerr2022self}
J.~Kerr, H.~Huang, A.~Wilcox, R.~Hoque, J.~Ichnowski, R.~Calandra, and K.~Goldberg, ``Self-supervised visuo-tactile pretraining to locate and follow garment features,'' \emph{arXiv preprint arXiv:2209.13042}, 2022.

\bibitem{lee2019making}
M.~A. Lee, Y.~Zhu, K.~Srinivasan, P.~Shah, S.~Savarese, L.~Fei-Fei, A.~Garg, and J.~Bohg, ``Making sense of vision and touch: Self-supervised learning of multimodal representations for contact-rich tasks,'' in \emph{2019 International conference on robotics and automation (ICRA)}.\hskip 1em plus 0.5em minus 0.4em\relax IEEE, 2019, pp. 8943--8950.

\bibitem{lin2019learning}
J.~Lin, R.~Calandra, and S.~Levine, ``Learning to identify object instances by touch: Tactile recognition via multimodal matching,'' in \emph{2019 International Conference on Robotics and Automation (ICRA)}.\hskip 1em plus 0.5em minus 0.4em\relax IEEE, 2019, pp. 3644--3650.

\bibitem{gao2023objectfolder}
R.~Gao, Y.~Dou, H.~Li, T.~Agarwal, J.~Bohg, Y.~Li, L.~Fei-Fei, and J.~Wu, ``The objectfolder benchmark: Multisensory learning with neural and real objects,'' in \emph{Proceedings of the IEEE/CVF Conference on Computer Vision and Pattern Recognition}, 2023, pp. 17\,276--17\,286.

\bibitem{T3}
L.~W. Jialiang~Zhao1, Yuxiang~Ma2 and E.~H. Adelson1, ``Transferable tactile transformers for representation learning across diverse sensors and tasks,'' \emph{arXiv preprint arXiv:2406.13640v1}, 2024.

\bibitem{wang2021multi}
L.~Wang and A.~v.~d. Oord, ``Multi-format contrastive learning of audio representations,'' \emph{arXiv preprint arXiv:2103.06508}, 2021.

\bibitem{rodriguez2024touch2touch}
\BIBentryALTinterwordspacing
S.~Rodriguez, Y.~Dou, M.~Oller, A.~Owens, and N.~Fazeli, ``Touch2touch: Cross-modal tactile generation for object manipulation,'' 2024. [Online]. Available: \url{https://arxiv.org/abs/2409.08269}
\BIBentrySTDinterwordspacing

\bibitem{yuan2017gelsight}
W.~Yuan, S.~Dong, and E.~Adelson, ``Gelsight: High-resolution robot tactile sensors for estimating geometry and force,'' \emph{Sensors}, vol.~17, p. 2762, 11 2017.

\bibitem{taylor2022gelslim}
I.~H. Taylor, S.~Dong, and A.~Rodriguez, ``Gelslim 3.0: High-resolution measurement of shape, force and slip in a compact tactile-sensing finger,'' in \emph{2022 International Conference on Robotics and Automation (ICRA)}.\hskip 1em plus 0.5em minus 0.4em\relax IEEE, 2022, pp. 10\,781--10\,787.

\bibitem{kuppuswamy2020soft}
N.~Kuppuswamy, A.~Alspach, A.~Uttamchandani, S.~Creasey, T.~Ikeda, and R.~Tedrake, ``Soft-bubble grippers for robust and perceptive manipulation,'' in \emph{2020 IEEE/RSJ International Conference on Intelligent Robots and Systems (IROS)}.\hskip 1em plus 0.5em minus 0.4em\relax IEEE, 2020, pp. 9917--9924.

\bibitem{digit_tactile_sensor}
\BIBentryALTinterwordspacing
``Digit tactile sensor - gelsight.'' [Online]. Available: \url{https://www.gelsight.com/product/digit-tactile-sensor/}
\BIBentrySTDinterwordspacing

\bibitem{fingerVision}
A.~Yamaguchi, ``Fingervision for tactile behaviors , manipulation , and haptic feedback teleoperation,'' 2018.

\bibitem{Do2022DenseTactOT}
W.~K. Do and M.~Kennedy, ``Densetact: Optical tactile sensor for dense shape reconstruction,'' \emph{2022 International Conference on Robotics and Automation (ICRA)}, pp. 6188--6194, 2022.

\bibitem{kuppuswamy2019fast}
N.~Kuppuswamy, A.~Castro, C.~Phillips-Grafflin, A.~Alspach, and R.~Tedrake, ``Fast model-based contact patch and pose estimation for highly deformable dense-geometry tactile sensors,'' \emph{IEEE Robotics and Automation Letters}, vol.~5, no.~2, pp. 1811--1818, 2019.

\bibitem{kim2022active}
S.~Kim and A.~Rodriguez, ``Active extrinsic contact sensing: Application to general peg-in-hole insertion,'' in \emph{2022 International Conference on Robotics and Automation (ICRA)}.\hskip 1em plus 0.5em minus 0.4em\relax IEEE, 2022, pp. 10\,241--10\,247.

\bibitem{suresh2023midastouch}
S.~Suresh, Z.~Si, S.~Anderson, M.~Kaess, and M.~Mukadam, ``Midastouch: Monte-carlo inference over distributions across sliding touch,'' in \emph{Conference on Robot Learning}.\hskip 1em plus 0.5em minus 0.4em\relax PMLR, 2023, pp. 319--331.

\bibitem{xu2023visual}
W.~Xu, Z.~Yu, H.~Xue, R.~Ye, S.~Yao, and C.~Lu, ``Visual-tactile sensing for in-hand object reconstruction,'' in \emph{Proceedings of the IEEE/CVF Conference on Computer Vision and Pattern Recognition}, 2023, pp. 8803--8812.

\bibitem{yamaguchi2016combining}
A.~Yamaguchi and C.~G. Atkeson, ``Combining finger vision and optical tactile sensing: Reducing and handling errors while cutting vegetables,'' in \emph{2016 IEEE-RAS 16th International Conference on Humanoid Robots (Humanoids)}.\hskip 1em plus 0.5em minus 0.4em\relax IEEE, 2016, pp. 1045--1051.

\bibitem{oller2023manipulation}
M.~Oller, M.~P. i~Lisbona, D.~Berenson, and N.~Fazeli, ``Manipulation via membranes: High-resolution and highly deformable tactile sensing and control,'' in \emph{Conference on Robot Learning}.\hskip 1em plus 0.5em minus 0.4em\relax PMLR, 2023, pp. 1850--1859.

\bibitem{ai2024robopack}
B.~Ai, S.~Tian, H.~Shi, Y.~Wang, C.~Tan, Y.~Li, and J.~Wu, ``Robopack: Learning tactile-informed dynamics models for dense packing,'' in \emph{ICRA 2024 Workshop on 3D Visual Representations for Robot Manipulation}.

\bibitem{chen2020simple}
T.~Chen, S.~Kornblith, M.~Norouzi, and G.~Hinton, ``A simple framework for contrastive learning of visual representations,'' in \emph{Proceedings of the 37th International Conference on Machine Learning}, ser. ICML'20.\hskip 1em plus 0.5em minus 0.4em\relax JMLR.org, 2020.

\bibitem{he2016deep}
K.~He, X.~Zhang, S.~Ren, and J.~Sun, ``Deep residual learning for image recognition,'' in \emph{Proceedings of the IEEE conference on computer vision and pattern recognition}, 2016, pp. 770--778.

\bibitem{resnet}
------, ``Deep residual learning for image recognition,'' in \emph{2016 IEEE Conference on Computer Vision and Pattern Recognition (CVPR)}, 2016, pp. 770--778.

\bibitem{imagenet}
J.~Deng, W.~Dong, R.~Socher, L.-J. Li, K.~Li, and L.~Fei-Fei, ``Imagenet: A large-scale hierarchical image database,'' in \emph{2009 IEEE Conference on Computer Vision and Pattern Recognition}, 2009, pp. 248--255.

\bibitem{tsne}
L.~Van~der Maaten and G.~Hinton, ``Visualizing data using t-sne.'' \emph{Journal of machine learning research}, vol.~9, no.~11, 2008.

\end{thebibliography}
